\pdfoutput=1

\documentclass[11pt]{article}

\usepackage{EMNLP2023}

\usepackage{times}
\usepackage{latexsym}

\usepackage[T1]{fontenc}

\usepackage[utf8]{inputenc}

\usepackage{microtype}

\usepackage{inconsolata}

\usepackage{amsmath}
\usepackage{mathtools}
\usepackage{amsfonts}
\usepackage{graphicx}
\usepackage{multicol}
\usepackage{multirow}
\usepackage{stmaryrd}
\usepackage{hyperref}
\usepackage{booktabs}
\usepackage{blindtext}
\usepackage{makecell}
\usepackage{comment}

\DeclareMathOperator*{\argmax}{arg\,max} 

\newcommand{\Thead}[1]{\textbf{\textsc{#1}}}
\newcommand{\frameworkname}{{\fontfamily{lmtt}\selectfont PRESQUE}\xspace}
\newcommand{\datasetname}{{\fontfamily{lmtt}\selectfont QuRe}\xspace}
\title{Pragmatic Reasoning Unlocks Quantifier Semantics for Foundation Models}

\author{
Yiyuan Li \qquad  Rakesh R. Menon \qquad  Sayan Ghosh \qquad  Shashank Srivastava\\
UNC Chapel Hill\\
\texttt{\{yiyuanli, rrmenon, sayghosh, ssrivastava\}@cs.unc.edu}
}

\begin{document}
\maketitle
\begin{abstract}
Generalized quantifiers (e.g., \textit{few, most}) are used to indicate the proportions predicates are satisfied (for example, \textit{some} apples are red). One way to interpret quantifier semantics is to explicitly bind these satisfactions with percentage scopes (e.g., \textit{30\%-40\%} of apples are red). This approach can be helpful for tasks like logic formalization and surface-form quantitative reasoning~\cite{Gordon2010QuantificationalSO, roy-etal-2015-reasoning}. However, it remains unclear if recent foundation models possess this ability, as they lack direct training signals. To explore this, we introduce \datasetname, a crowd-sourced dataset of human-annotated generalized quantifiers in Wikipedia sentences featuring percentage-equipped predicates. We explore quantifier comprehension in language models using \frameworkname, a framework that combines natural language inference and the Rational Speech Acts framework. Experimental results on the HVD dataset and \datasetname illustrate that \frameworkname, employing pragmatic reasoning, performs 20\% better than a literal reasoning baseline when predicting quantifier percentage scopes, with no additional training required\footnote{Code: \url{https://github.com/Nativeatom/PRESQUE}}.
\end{abstract}

\section{Introduction}
Generalized quantifiers~\citep{Mostowski1957OnAG} are used to express relations between subsets of concepts or entities. For instance, the quantifier `\textit{some}' in the statement `\textit{some apples are red}' indicates that at least one apple is red. Quantifiers, being inherently fuzzy, are prevalent in both real-world communication and natural language processing (NLP) benchmarks \citep{joshi-etal-2020-taxinli}. Consequently, developing a formal framework for understanding quantifier semantics is essential to enhance the language understanding capabilities of NLP systems, particularly in facilitating natural human-AI language-based interactions, such as in human-robot collaborative tasks \citep{human_robot_collaborative_task}.

In this work we present \frameworkname~(\textbf{P}ragmatic \textbf{RE}asoning for \textbf{S}emantics of \textbf{QU}antifi\textbf{E}rs), a framework to model the semantics of quantifiers for text-based foundation models, such as BERT~\citep{devlin-etal-2019-bert} and RoBERTa~\citep{Liu2019RoBERTaAR}, through the lens of pragmatic reasoning. While foundation models have shown impressive performance on various text-based tasks \citep{Bommasani2021OnTO, wei2022emergent}, their ability to infer the semantic meanings of generalized quantifiers remains relatively unexplored.

In \frameworkname, we represent quantifier semantics in terms of percentage scopes, which indicate the proportion of cases where the associated predicate holds true. For example, in the sentence `some apples are red', the quantifier `some' could be associated with a percentage scope of 30-40\%, indicating that 30-40\% of all apples are red. Our framework consists of two components: (1) a natural language inference (NLI) component \citep{bowman-etal-2015-large} that models sentence-level semantics between a sentence containing a quantifier word and another sentence containing a percentage value, and (2) a rational speech act (RSA) component \citep{rsa/science.1218633} for pragmatic reasoning. Using these components, \frameworkname takes a sentence with a quantifier as input and outputs the corresponding percentage scope (further discussed in Section~\ref{sec: Rational Speech Act}).

Ambiguity, as highlighted by \citet{Piantadosi2012TheCF}, is beneficial for efficient communication via language. %
Since the percentage values of quantifiers are not universally defined, humans often need to infer the exact percentage value, which is not explicitly conveyed in the utterance \citep{Horowitz2018TheTW}. 
Furthermore, the interpretation of quantifier semantics can be influenced by linguistic and social cues \citep{Bergen2016PragmaticRT}. The pragmatic theory proposed by \citet{LogicandConversation} emphasizes the role of communicative goals in interpreting the semantic meaning of natural language expressions, simplifying the required semantic theories \citep{Bergen2016PragmaticRT}. Lastly, NLI models are also shown to struggle with ambiguous premises~\citep{thukral-etal-2021-probing}, quantifier inference~\cite{Richardson_Hu_Moss_Sabharwal_2020, joshi-etal-2020-taxinli}, and quantitative reasoning~\cite{naik-etal-2018-stress, ravichander-etal-2019-equate}, making a direct literal interpretation of generalized quantifiers less reliable. 

To address these challenges, \frameworkname employs RSA, a Bayesian framework that follows a Gricean approach for modeling communication by reasoning about agent states \citep{LogicandConversation}. \frameworkname incorporates a literal speaker role, based on a foundation model fine-tuned on NLI datasets, and a pragmatic listener role, computed using Bayesian rules, to reason between the quantifier space and the space of percentage values.

Existing datasets like HVD~\citep{2015FromCT} and GQNLI~\cite{cui-etal-2022-generalized-quantifiers} that investigate quantifier semantics either lack gold annotations of the percentage scopes for interpreting quantifier semantics, or are based on artificial setups using a small number of countable objects~\cite{pezzelle-etal-2018-comparatives}. Such fictional settings are not generalizable to broader and more complex real-world settings (e.g. describing concepts about a population using quantifiers). 
For a fair evaluation of the quantifier understanding capabilities acquired by foundation models through their pre-training, we should evaluate these models using text of similar style and content as the pre-training corpora. 
To address the aforementioned issues with current evaluation corpora for quantifier understanding, we crowd-source a dataset, \datasetname (\textbf{Qu}antifier \textbf{Re}asoning), which contains sentences containing quantifiers paired with annotations for quantifier semantics in terms of percentage scopes. Additionally, we characterize the ease of making quantifier predictions for different sentences in \datasetname.

Using \frameworkname to evaluate the quantifier reasoning ability of foundation models on \datasetname, we observe a 20\% span-based F1 boost over the literal listener baseline at all specificity levels (Section~\ref{sec: model selection}). 

Our experiments highlight the improved quantifier understanding of foundation models when approached from a pragmatic perspective rather than relying on direct interpretation using textual understanding frameworks like NLI. Although our framework does not explicitly model mathematical concepts, it is noteworthy that the mean strengths of quantifiers in foundation models, as revealed by \frameworkname, echo observations of quantifier hierarchies from previous works~\cite{Solt2016OnMA, srivastava2018zero} that involve strong human priors, and findings from \citet{PEZZELLE2018117}, who associates quantifier usage with the counting behavior of human beings.

In summary, our contributions are two-fold: we develop \frameworkname based on pragmatic reasoning and NLI, and we crowd-source a dataset \datasetname to support quantifier understanding investigation of foundation models. Our results on HVD and \datasetname demonstrate that foundation models equipped with pragmatic reasoning (\frameworkname) can perform quantifier reasoning similar to humans.

\section{Quantifier Semantics Understanding through RSA}
\label{sec: Rational Speech Act}
We frame the task of quantifier understanding as the prediction of the percentage scope (e.g., 30\%-50\%) given a quantified sentence $\Tilde{S}_{q}$ (e.g., \textit{Some apples are red.}). Specifically, given an interval width $\beta$, we divide the percentage spectrum between 0 and 1 into evenly spaced intervals, denoted as $\mathcal{W}_{\beta} = \{p_{i}\}$ (e.g., $\mathcal{W}_{\beta=0.05} = \{0, 5\%, 10\%, ..., 95\%, 100\%\}$).  The goal of a quantifier understanding model is to determine the percentage range in $\mathcal{W}_{\beta}$ where the associated predicate holds true (e.g., the proportion of red apples among all apples, 30\%-50\%).

To interpret quantifiers as percentage scopes, we develop \frameworkname, a framework that adopts the rational speech act (RSA) framework, with natural language inference (NLI) as the backbone for text understanding. The RSA framework consists of a speaker and a listener, where the listener infers the world state from the speaker's utterance by modeling the speaker's state of mind \citep{GOODMAN2016818}. In \frameworkname, the world state corresponds to percentage values of predicates, while utterances correspond to quantifiers used with those predicates. 

Given a premise $\Tilde{p}$ (e.g., \textit{Some apples are red.}) with quantifier $q$ (\textit{some}) and a hypothesis $\Tilde{h}$ (e.g., \textit{30\% apples are red.}) with a percentage value $p$ (\textit{30\%}), we use the entailment score between the premise and hypothesis, obtained from an NLI model, to define the literal listener $\mathrm{L}_{0}$: 

\begin{align}
    \mathrm{L}_{0}(p|q) \propto \text{Entailment}(\Tilde{p}, \Tilde{h}) \label{eq: literal listener}
\end{align}

The pragmatic listener $L_{1}$, in the \frameworkname framework, interprets the semantics of the quantifier word as:
\begin{align}
    \mathrm{L}_{1}(p|q) \propto \mathrm{S}_{0}(q|p)\mathbb{P}(p) \label{eq:pragmatic listener}
\end{align}

Here, $\mathrm{S}_{0}$ represents a literal speaker that maps the semantics of percentage values to quantifier words. Practically, we model the speaker by swapping the premise and hypothesis in $\mathrm{L}_{0}$:
\begin{align}
    \mathrm{S}_{0}(q|p) \propto \text{Entailment}(\Tilde{h}, \Tilde{p}) \label{eq: literal speaker}
\end{align}

The prior $P(p)$ in Eq.~\ref{eq:pragmatic listener} can be expanded as:
\begin{align}
    \mathbb{P}(p) = \sum_{q \in \mathcal{U}} \mathbb{P}(p|q)\mathbb{P}(q) \label{eq:bayesian prior} 
\end{align}
Here, $\mathbb{P}(p|q)$ is computed similarly to $\mathrm{L}_0$, and $\mathbb{P}(q)$ represents the word frequency of $q$, obtained from the WORDFREQ dataset \citep{robyn_speer_2022_7199437}. 

\section{\datasetname: Quantifier Reasoning Dataset}
\label{sec: dataset collection}

Existing datasets for quantifier understanding like HVD~\cite{2015FromCT} are comprised of triples of the form $\langle$concept, feature, quantifier$\rangle$ (e.g. $\langle$\textit{shrimp}, \textit{is\_white}, \textit{some}$\rangle$) that denote how often a `feature' is satisfied for a `concept'. Notably, these datasets do not provide annotated percentage scopes that can be used to decipher the semantics of the quantifiers, i.e., how often (in numerical terms) the `feature' is satisfied for the `concept' in the real world, and the supporting documents (e.g. knowledge-bases in Wikipedia or any publicly available corpus) about the percentage scope of those triples are not easily accessible. Therefore, the judgments are based on subjective observation and experience (e.g. the proportion of white shrimps.) and are hence inaccurate. To address this shortcoming in available resources for quantifier understanding, we contribute a dataset, \datasetname, for evaluating the quantifier understanding capabilities of language models. \datasetname consists of sentences (from Wikipedia) containing percentage mentions annotated with the quantifiers. 

Table~\ref{tab: QuRe dataset examples shortened} presents examples from  \datasetname. Of note, in addition to the quantifier annotation and percentage scopes, for each example in \datasetname, we also provide specificity as additional metadata. Specificity measures the difficulty of deciphering the percentage scope of the quantifier from the sentence excluding the quantifier (i.e., if someone can deduce the percentage scope of a quantifier fully/partially from the sentence contents when the quantifier is absent; more details are provided later in Stage 4). The annotations in \datasetname are obtained through a mix of crowd-sourcing (using Amazon Mechanical Turk) and assistance from gpt-3.5-turbo~\cite{openai2022chatgpt}. We describe the annotation procedure in more detail below.

\begin{table} 
    \footnotesize
    \centering
    \begin{tabular}{p{3.3cm}|p{3.4cm}}
    \toprule
        \Thead{Wikipedia Sentence} & \Thead{[Specificity, Expression] 
  \datasetname{} Sentence}\\
        \midrule
        \footnotesize{Squirrel Hill North's population is 75\% White, 17\% Asian, \underline{4\%} Hispanic, and 3\% Black.} & \footnotesize{\textbf{[Partial, $0.04$]} Squirrel Hill North's population is 75\% White, 17\% Asian, \underline{few} Hispanic, and 3\% Black.}\\
        \midrule 
        \footnotesize{Coconut milk contains 5\% to 20\% fat, while coconut cream contains \underline{around 20\% to 50\%} fat.}.& \footnotesize{\textbf{[Indeterminable, $0.2-0.5$]} Coconut milk contains 5\% to 20\% fat, while coconut cream contains \underline{moderate} fat.}\\
        \bottomrule
    \end{tabular}
    \caption{Examples of \datasetname, with target percentage mention and the quantifier \underline{underlined}. The headers of the \datasetname{} also provide information about specificity and percentage expression generated. More examples are included in Appendix~\ref{sec: QuRe metadata example}. 
    }
    \label{tab: QuRe dataset examples shortened}
\end{table}

\paragraph{Stage 1: Wikipedia sentences collection} We utilize the $\langle$concept, feature, quantifier$\rangle$ triples from    the HVD dataset and convert them into sentences programmatically through templates (e.g. $\langle$\textit{shrimp}, \textit{is\_white}, \textit{some}$\rangle$ $\rightarrow$ `some shrimps are white'). We then link these sentences to the most related Wikipedia entities\footnote{Each Wikipedia entity is the title of a Wikipedia article.} using an off-the-shelf library\footnote{\url{https://pypi.org/project/wikipedia/}}. For example, the related Wikipedia entities for the running example would be \textit{\{Shrimp, Prawn, Indian prawn, etc.\}}. In practice, we find this setting links to more diverse entities than simply linking the concepts. We then use regular expressions to extract around 5,000 candidate sentences containing percentage values from the Wikipedia pages of these entities. For example, \textit{`Among the prawn species entering the field F. indicus constitute \underline{around 36\%–43\%}.'} is one such sentence from the Wikipedia page of the entity \textit{Indian prawn}, with the percentage mention \underline{underlined}. 

\paragraph{Stage 2: Sentence Filtering} The candidate sentences are further filtered based on two criteria: (1) the length of the sentences should be between 10 and 40 tokens (space-tokenized), and (2) the percentage mentioned in the sentence should not indicate a comparative change (e.g., \textit{`increased by 20\%'}). To identify whether the sentence describes a comparative change, we used regular expressions. However, capturing all possible variations of describing such comparative changes through regular expressions is cumbersome. Hence we employ GPT-3.5-turbo 
to annotate sentences that contain comparative changes. To validate the efficacy of GPT-3.5-turbo, we manually annotate a held-out set of 50 sentences based on our aforementioned filtering criteria. On this held-out set GPT-3.5-turbo achieves 0.76 F1. More details on the annotation usage of GPT-3.5-turbo in this stage are included in  Appendix~\ref{sec: instruction-quantifier-filtering}. The filtered sentences are then paired up with all percentage mentions in the sentence and manually validated by the authors. Around half of the percentage mentions were deemed inappropriate for the quantifier understanding task and removed. We include examples, metadata, and the instruction used in Appendix~\ref{sec: QuRe metadata example} and ~\ref{sec: sentence topic generation}. 

\paragraph{Stage 3: Percentage Expression Generation}
In many Wikipedia sentences, the percentage value is surrounded by texts like \textit{around}, \textit{less than}, \textit{more than}, etc. to denote a percentage scope rather than the individual percentage value or a percentage range. We use GPT-3.5-turbo to obtain those percentage scopes and the instruction is included in Appendix~\ref{sec: instruction-mathematical expression}. The variations that we capture in this stage to obtain the percentage scopes are mentioned in Table~\ref{tab: percentage scope generation operators}. 

\begin{table}[!ht]
    \small
    \centering
    \begin{tabular}{c|l}
    \toprule
        \Thead{Op.} & \Thead{Percentage Mention: Expression}\\
        \midrule
        None & 89\%: $0.89$\\
        \midrule
        $>$  & over 93\%: $>0.93$\\
        \midrule
        $>=$  & at least 45\%: $>=0.45$\\
        \midrule
        $<$  & less than 1\%: $<0.01$\\
        \midrule
        $<=$  & not exceeding 19\%: $<=0.19$\\
        \midrule
        $-$  & between 24\% and 40\%: $0.24-0.4$\\
        \midrule
        $\sim$ & about 98\%: $\sim0.98$\\
        \bottomrule
    \end{tabular}
    \caption{Operators (\Thead{Op.}) in percentage expression generation and examples.}
    \label{tab: percentage scope generation operators}
\end{table}

\paragraph{Stage 4: Quantifier and Specificity Annotation} 
We design two human annotation tasks. The first task is rephrasing a sentence, $\Tilde{S}_{p}$, with a target percentage mention (e.g. `\textit{around 36\%-43\%}' of `\textit{...the field F. indicus constitute \underline{around 36\%–43\%}}' in the previous example) to $\Tilde{S}_{q}$ with minimal semantic changes using a quantifier from $\mathcal{U}$ (e.g. \textit{Among the prawn species entering the field, F. indicus constitute \underline{a large amount}.}). This step ensures the semantics of the quantifier used in $\Tilde{S}_{q}$ is associated to the percentage scope in $\Tilde{S}_{p}$.

In the second task, we measure specificity, or the difficulty of specifying the target percentage scope from $\Tilde{S}_{q}$ without the quantifier $q$ (e.g. removing \textit{a large amount} from the previous $\Tilde{S}_{q}$). In our study, we discretize the specificity values into three distinct levels of difficulty: \textit{full/partial/indeterminable}. \textit{Full} means the target percentage scope can be fully determined by other contents in the sentence, like \textit{One in ten} for \textit{10\%}; \textit{partial} means the percentage scope can be narrowed but not determined by the contents (e.g. an incomplete percentage breakdown), and \textit{indeterminable} means there is no information in the content of the sentence to determine the percentage scope.  This task aims to gauge the extent of information that the context contributes to the determination of the quantifier's percentage scope. For example, the specificity of the previous $\Tilde{S}_{q}$ about prawn would be \textit{indeterminable} since the rest of the sentence after removing \textit{a large amount} does not provide information to determine the percentage scope of \textit{large}. But with additional contents like `\textit{... constitute a large amount (around one-third).}', the specificity level would become \textit{partially}. More examples are included in Appendix~\ref{sec: QuRe metadata example}. 

We use majority voting (amongst three annotations) to choose the final annotated quantifier among all annotations for each sentence. 
The instruction used, examples, and example annotation interface are included in Appendix~\ref{sec: annotation task interface}. The set of quantifiers to select from is $\mathcal{U}$ = \{\textit{all, generally, most, usually, some, likely, few, little, occasionally, none, seldom, tiny, small, moderate, large}\}, which largely comes from~\citet{srivastava2018zero}, and is slightly extended based on preliminary annotator feedback. We leave the choice of nouns that are attached to adjective quantifiers (e.g. \textit{amount} in \textit{small amount}), like \textit{small} and \textit{large}, in sentences to the annotators.

\begin{figure}[!t]
    \centering
    \includegraphics[width=1\linewidth]{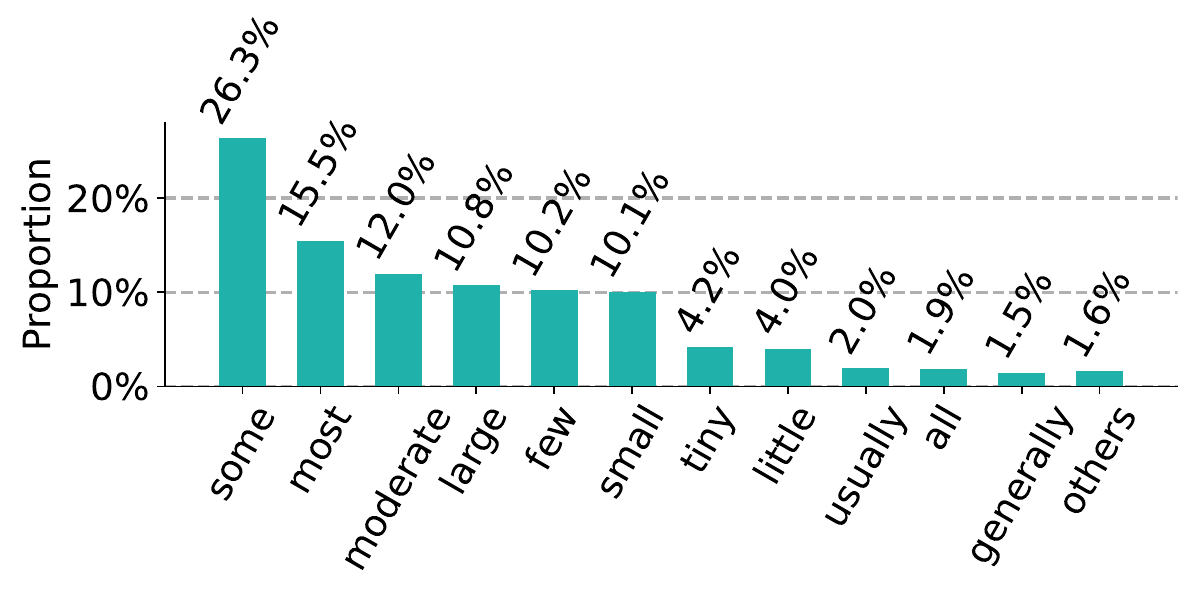}
    \caption{Distribution of quantifiers in \datasetname{}. Quantifiers with less than 1\% frequencies (\textit{likely, seldom, occasionally, none}) are merged into \textit{others}. \textit{Some}, \textit{most} and \textit{moderate} are the most frequent quantifiers in \datasetname{}. 
    }
    \label{fig: bar chart quantifier distribution in QuRe}
\end{figure}

\begin{figure}[!ht]
    \centering
    \includegraphics[width=1.0\linewidth]{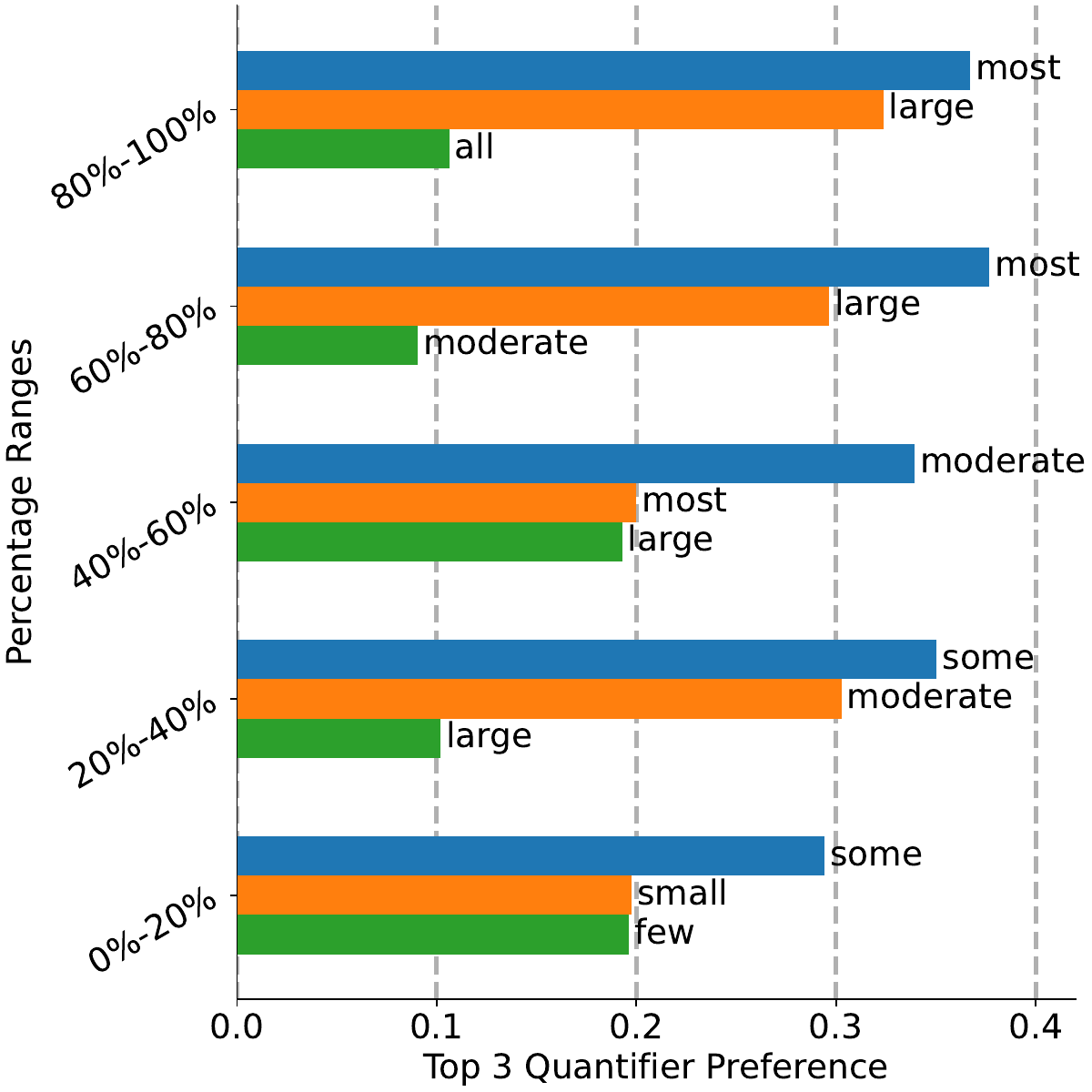}
    \caption{Quantifier preferences in difference percentage scopes, e.g., \textit{most} is chosen around 35\% of the times if the percentage mentioned lies in 60\%-100\%.}
    \label{fig: quantifier preference by bucket}
\end{figure}

\paragraph{Statistics} We have collected 744 $\Tilde{S}_{q}$ sentences, of which 47\% and 17\% contain no and one percentage mention respectively and others contain more than one percentage mention. The average sentence length is 26.3 tokens. Each sentence is annotated by 3 annotators. The Fleiss' Kappa for quantifier choices and specificity are 0.37 and 0.80, meaning fair agreement in quantifier choices and substantial agreement in specificity levels. The distribution of quantifiers in \datasetname is shown in Figure~\ref{fig: bar chart quantifier distribution in QuRe}, where \textit{some} is used in over 25\% of sentences, followed by \textit{most}, \textit{moderate}, \textit{large} and \textit{few}. The most popular quantifiers for different percentage scopes are shown in Figure~\ref{fig: quantifier preference by bucket}, where \textit{some} is preferred in over 30\% of the cases with target percentage values lower than 40\%, and \textit{most} is selected in over 40\% of the cases with target percentage value greater than 60\%. Overall, 17\% of the sentences have target percentages fully specified, 32\% partially specified, and 50\% are indeterminable. We include examples across different specificity levels in Appendix~\ref{sec: QuRe metadata example}.

We also include the average strength of quantifiers under different grounding configurations in Table~\ref{tab: mean quantifier strength}.\footnote{For the definition of $g$ and $w$, please refer to Appendix \ref{sec: percentage scope grounding details}.} We can see that the mean strengths are stable among configurations, and show interesting hierarchies: \textit{all} (0.88) is higher than \textit{generally} (0.73), and \textit{generally} is higher than \textit{most} (0.69). These patterns closely align with previous manual strength assignments like ~\citet{srivastava2018zero}, and \citet{testoni-etal-2019-quantifiers}'s quantifier collection from multimodal simulations. It also echoes \citet{Solt2016OnMA}'s finding that the strength of \textit{most} is higher than \textit{more than half}.

\begin{table}[]
    \small
    \centering
    \begin{tabular}{c|c|c}
    \toprule
    Quantifier & $g=0.01$ $w=1$ &  $g=0.01$ $w=4$\\
    \midrule
    all & 0.885 $\pm$ 0.087 &  0.892 $\pm$ 0.085 \\
    \midrule
    generally & 0.730 $\pm$ 0.205 & 0.708 $\pm$ 0.212\\
    \midrule
    usually & 0.686 $\pm$ 0.249 &  0.674 $\pm$ 0.242\\
    \midrule
    most & 0.687 $\pm$ 0.193 &  0.693 $\pm$ 0.195\\
    \midrule
    large & 0.624 $\pm$ 0.217 & 0.628 $\pm$ 0.223\\
    \midrule
    likely & 0.473 $\pm$ 0.287 &  0.504 $\pm$ 0.266\\
    \midrule
    moderate & 0.369 $\pm$ 0.154 &   0.372 $\pm$ 0.156\\
    \midrule
    some & 0.225 $\pm$ 0.185 &  0.218 $\pm$ 0.182\\
    \midrule
    small & 0.183 $\pm$ 0.184 &  0.172 $\pm$ 0.172\\
    \midrule
    occasionally & 0.119 $\pm$ 0.037 & 0.124 $\pm$ 0.037\\
    \midrule
    seldom & 0.112 $\pm$ 0.117 & 0.093 $\pm$ 0.106\\
    \midrule
    little & 0.104 $\pm$ 0.109 & 0.117 $\pm$ 0.135\\
    \midrule
    few & 0.074 $\pm$ 0.087 & 0.081 $\pm$ 0.098\\
    \midrule
    tiny & 0.024 $\pm$ 0.048 &  0.031 $\pm$ 0.046\\
    \midrule
    none & 0.004 $\pm$ 0.007 &  0.004 $\pm$ 0.007\\
    \bottomrule
    \end{tabular}
    \caption{Average strengths of quantifiers in all annotations of \datasetname under different grounding configurations. The average strengths are stable with different window sizes.}
    \label{tab: mean quantifier strength}
\end{table}

\section{Experimental setup}
\label{sec: setup}
For the experiment in HVD, we compute $\mathrm{L}(p|q)$ for \frameworkname and $\mathrm{L}_{0}$ among different foundation models and compare them with human interpretations.
In \datasetname, with the target percentage given in Section~\ref{sec: dataset collection}, we generate the percentage scope that $\Tilde{S}_{q}$ satisfies. All percentage choices are selected from $\mathcal{W}_{\beta}$, and experiments are run without training.

\paragraph{Percentage Scope Generation} 
With specific granularity $g$ and window size $w$ for the operators, a percentage expression in Section~\ref{sec: dataset collection} is converted into a golden scope 
$\{p_{\text{min}}, p_{\text{max}}\} \in \mathcal{W}_{\beta} \quad (p_{\text{min}}\leq p_{\text{max}})$. For example, if $\beta=0.05$, $g=0.01$ and $w=2$, the golden scope of $\sim 0.59$ is $[0.55, 0.65]$. The full generation rules are in Appendix~\ref{sec: percentage scope grounding details}.

\begin{figure*}[!ht]
    \centering
    \includegraphics[width=1\linewidth]{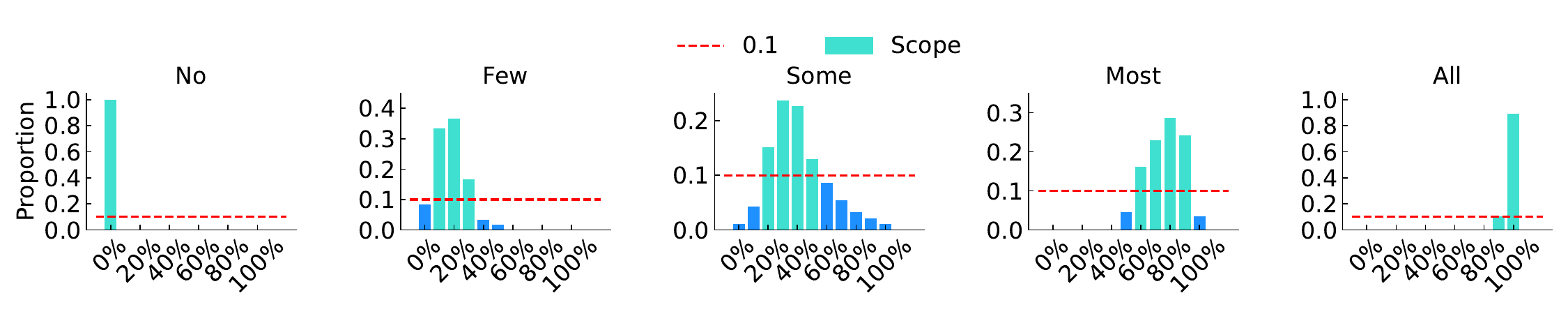}
    \includegraphics[width=1\linewidth]{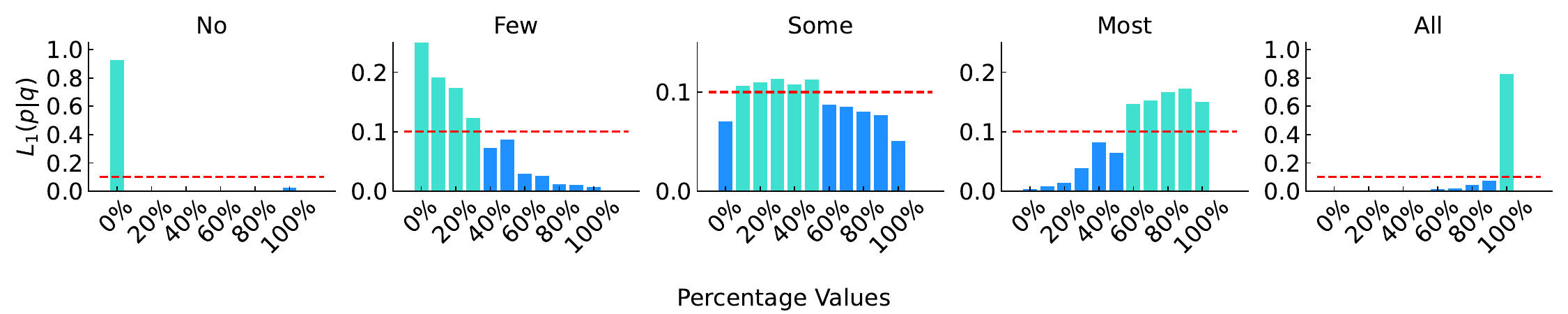}
    \caption{Human interpretations (upper) from 25 annotators and \frameworkname scores ($\mathrm{L}_{1}$) from RoBERTa-large (bottom) of quantifier percentage scopes in HVD. The cyan bars indicate percentage values chosen by more than 10\% of the annotators (red line) and, therefore could serve as approximate percentage scopes. For example, 10\%-30\% for \textit{few} in human interpretations. The threshold is only used for illustration and not in experiments.}
    \label{fig: human-quantifier2percentage_range}
\end{figure*} 
\paragraph{Evaluation Metrics} For HVD, given a listener $\mathrm{L}_{\text{M}}$ based on an NLI model M. $\mathrm{L}_{\text{M}}(p|q)$ is computed by averaging the entailment scores over all $\Tilde{S}_{q}$s for all $p$ values in $\mathcal{W}_{\beta}$ and normalize them to be a distribution. We can then compute the cross entropy between the human interpretation of quantifiers $\mathbb{P}_{\text{h}}$ from Section~\ref{sec: human evaluation L1} and $\mathrm{L}_{\text{M}}(p|q)$ to measure the similarity of quantifier interpretation between humans and M.

\begin{align}
    \text{CrossEntropy} & = -\sum_{q \in \mathcal{U}}\sum_{p \in \mathcal{W}_{\beta}}\frac{\mathbb{P}_{\text{h}}(p|q)\log \mathrm{L}_{\text{M}}(p|q)}{{|\mathcal{U}|}} \nonumber
\end{align}
For $\Tilde{S}_{q}$ in \datasetname, we compute the following metrics,

\begin{flalign}
    &\text{HIT@1} = \mathbb{I}[\argmax_{p \in \mathcal{W}_{\beta}}\mathrm{L}(p|q) \in s_{\text{golden}}] && \nonumber \\
    & \text{MRR} =  \makecell{\beta/ (\text{B}_{m} \cdot \sum_{p' \in s_{\text{golden}}}\text{Rank}_{p'})} && \nonumber\\
    &\text{CrossEntropy} = \makecell{-\sum_{p' \in s_{\text{golden}}}\log \mathbb{P}(p'|q)} && \nonumber \\
    &\text{where} \quad \mathbb{P}(p'|q)) = \makecell{\mathrm{L}(p'|q) / \sum_{p}\mathrm{L}(p|q)} && \nonumber\\
    &\text{where} \quad \text{B}_{m} =  p_{\text{max}} - p_{\text{min}} + \beta, \quad p' \in \mathcal{W}_{\beta} \nonumber
\end{flalign}

where $\mathbb{I}(\cdot)$ is an indicator, $s_{\text{golden}}$ is the gold scope $[p_{\text{min}}, p_{\text{max}}]$, and $\text{Rank}_{p'}$ is the rank of $p'$ in $\mathcal{W}_{\beta}$ by $\mathrm{L}(p|q)$. HIT@1 measures whether the top inference percentage lies in the gold scope. MRR and cross entropy measure the average rank and confidence of the gold scope. \textcolor{black}{We also compute the span-based F1 score between the gold scope and the primary scope (Section~\ref{sec: model selection}) of the top K predictions (F1@K) under $\mathcal{W}_{\beta}$, which is used in question answering~\cite{rajpurkar-etal-2016-squad}.} The metrics are averaged over the entire dataset, and  $\mathrm{L}(p|q)$ is computed through Eq.~\ref{eq: literal listener} - Eq.~\ref{eq:bayesian prior}. 

\section{Experiments and Results}
\label{sec: Experiments and Results}

We perform experiments to determine the percentage scope of quantifiers on two datasets: the HVD dataset, which includes predicates annotated with quantifiers but lacks percentage scope annotations, and the \datasetname dataset, which provides annotations for both quantifiers and percentage scopes. As a baseline, we use the literal listener, $\mathrm{L}_{0}$.

\subsection{Human Interpretation of Quantifiers}
\label{sec: human evaluation L1}
To quantitatively assess the similarity between quantifier understanding between foundation models and humans, we first collect interpretations $\mathbb{P}_{h}(p|q)$ from human participants. For this, we employ 25 Mechanical Turk annotators who are tasked with providing the percentage scope of quantifiers. To guide them, we provide an instruction (see Appendix~\ref{sec: human evaluation instruction} for details) and present them with five questions. Each question requires the annotator to indicate the strength of a given quantifier word in $\mathcal{U}$ by providing either a percentage scope or a single percentage value in $\mathcal{W}_{\beta=0.1}$, without resorting to online searching. The distribution of the annotators' choices, as shown in Figure~\ref{fig: human-quantifier2percentage_range}, reveals that humans interpret different percentage peaks for different quantifier words. The percentage scope of \textit{few}, \textit{some}, \textit{most} indicated by the selection ratio of more than 10\% are larger than those of \textit{no} and \textit{all}. Meanwhile, the percentage scope of \textit{some} is leaning to \textit{few} rather than \textit{most}, where \textit{few} and \textit{most} have little scope overlap. 

\subsection{NLI Model's Interpretation of Quantifiers}
\label{sec: model selection}
We evaluate the quantifier reasoning ability of the \texttt{`large'} (or \texttt{`xxlarge'}) variants of ALBERT~\citep{Lan2020ALBERT:}, XLNet~\citep{NEURIPS2019_xlnet}, BART~\citep{lewis-etal-2020-bart} and RoBERTa~\citep{Liu2019RoBERTaAR} that are fine-tuned on the NLI tasks (using Adversarial NLI~\citep{nie-etal-2020-adversarial}, SNLI~\citep{bowman-etal-2015-large} MNLI~\citep{MNLI}, and NLI-style FEVER~\citep{nie2019combining} datasets).\footnote{In preliminary experiments, we found that foundation models without NLI fine-tuning performed worse on the quantifier prediction task.}

\begin{figure*}[!ht]
    \centering
    \includegraphics[width=\linewidth]{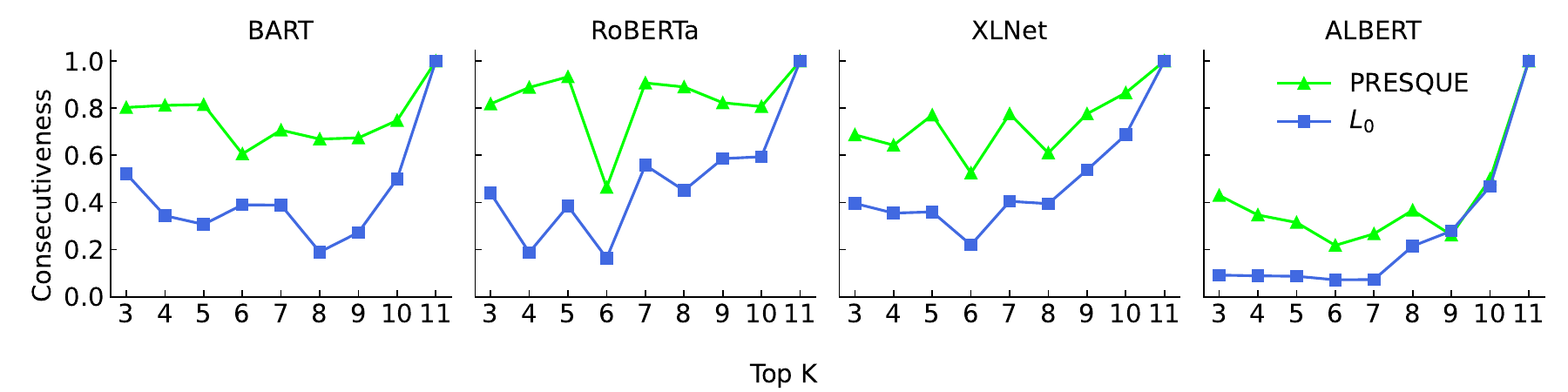}
    \caption{The ratio of top K percentage values from \frameworkname (lime) 
 and $\mathrm{L}_{0}$ (blue) that can form a single consecutive scope in HVD. \frameworkname has higher consecutiveness than $\mathrm{L}_{0}$ among all models.
 }
    \label{fig: consecutiveness}
\end{figure*}

\begin{table}[!ht]
    \small
    \centering
    \begin{tabular}{l|c|c}
    \toprule
        \multirow{2}{*}{\Thead{Base Model}(\Thead{\#Param.})} & \multicolumn{2}{|c}{\Thead{CrossEntropy}$\downarrow$} \\
        \cmidrule{2-3}
        & $\mathrm{L}_{0}$ & \frameworkname\\
        \midrule
        \small{ALBERT (222M)} & 1.76 & 1.48\\
        \small{XLNet (361M)} & \textbf{1.64} & 1.35\\
        \small{BART (407M)} & 1.89 & 1.32\\
        \midrule
        \small{RoBERTa (355M)} & 1.69 & \textbf{1.29}\\
        \bottomrule
    \end{tabular}
    \caption{Comparison of different NLI models in HVD with $L_{0}$ being the baseline of using NLI models for direct interpretation and \frameworkname is the pragmatic-based interpretation. \frameworkname is better than $L_{0}$ and RoBERTa-large has best cross entropy in \frameworkname.}
    \label{tab: nli model comparison}
\end{table}

\begin{table}[]
    \small
    \centering
    \begin{tabular}{p{3cm}|l|p{1cm}}
    \toprule
        \Thead{Sentence} &\Thead{Scope} & \Thead{Pref.} \\
        \midrule
        \multirow{2}{*}{\multirowcell{2}[0ex][l]{\underline{No} ostriches are\\ strange looking.}} & $\mathrm{L}_{0}$: 0\%-40\% & 0.34 \\
        \cmidrule(lr){2-3}
         & $\mathrm{L}_{1}$: 0\%-10\% & \textcolor{blue}{\textbf{0.66}} \\
        \midrule
        \multirow{2}{*}{\multirowcell{2}[0ex][l]{\underline{Few} tomatoes\\ are green.}} & $\mathrm{L}_{0}$: 0\% & 0.12 \\
        \cmidrule(lr){2-3}
         & $\mathrm{L}_{1}$: 0\%-30\% & \textcolor{blue}{\textbf{0.88}} \\
        \midrule
        \multirow{2}{*}{\multirowcell{2}[0ex][l]{\underline{Some} kites are\\ made of plastic.}} & $\mathrm{L}_{0}$: 80\%-100\% & 0.38 \\
        \cmidrule(lr){2-3}
         & $\mathrm{L}_{1}$: 10\%-50\% & \textcolor{blue}{\textbf{0.62}} \\
        \midrule
        \multirow{2}{*}{\multirowcell{2}[0ex][l]{\underline{Most} owls live in forests.}} & $\mathrm{L}_{0}$: 80\%-100\% & \textcolor{green}{\textbf{0.66}} \\
        \cmidrule(lr){2-3}
         & $\mathrm{L}_{1}$: 60\%-100\% & \textcolor{black}{0.34} \\
        \midrule
        \multirow{2}{*}{\multirowcell{2}[0ex][l]{\underline{All} gates are\\ used for enclosing.}} & $\mathrm{L}_{0}$: 60\%-100\% & 0.22 \\
        \cmidrule(lr){2-3}
        & $\mathrm{L}_{1}$: 70\%-100\% & \textcolor{blue}{\textbf{0.78}} \\
        \bottomrule
    \end{tabular}
    \caption{Examples of percentage preferences between \frameworkname ($\mathrm{L}_{1}$) and $\mathrm{L}_{0}$ in HVD. The primary scope (Scope) is a scope with the maximum subarray sum of $L(p|q)$ within top K inference values, which stands for the most confident percentage scope of the model. The human preference (Pref.) is the ratio of scopes preferred by the human annotators. Green and blue represent preference to $\mathrm{L}_{0}$ and \frameworkname, respectively.}
    \label{tab:listener preference example}
\end{table}

The comparison of quantifier understanding using \frameworkname and $\mathrm{L}_{0}$ is presented in Table~\ref{tab: nli model comparison}. The results show that \frameworkname achieves lower cross entropies compared to $\mathrm{L}_{0}$. Among the NLI models, RoBERTa performs the best within the \frameworkname framework, and therefore, it is chosen as the primary model for subsequent experiments. The $\mathrm{L}(p|q)$ scores of \frameworkname from RoBERTa, which are used to represent the model's interpretation of the percentage scopes of different quantifiers, are displayed in the lower half of Figure~\ref{fig: human-quantifier2percentage_range}. In general, different quantifier words exhibit distinct percentage distributions. Similar to Section~\ref{sec: human evaluation L1}, the scopes of \textit{few}, \textit{some}, and \textit{most} can be approximated as \textit{0\% - 30\%}, \textit{10\% - 50\%}, and \textit{60\% - 100\%}, respectively, with a cutoff criteria $\mathrm{L}(p|q) \geq 0.1$. These ranges align closely with the scopes determined by human evaluation (upper half of Figure \ref{fig: human-quantifier2percentage_range}). 
Further, the $\mathrm{L}(p|q)$ scores change in a smooth way as the percentages increase or decrease within the regions where $\mathrm{L}(p|q) \geq 0.1$. This suggests that the model can understand percentage values quite well.

Next, we compare the results of \frameworkname with that of a literal listener baseline, $\mathrm{L}_{0}$ (Equation~\ref{eq: literal listener}). As the percentage scope of a quantifier is measured by the consecutive percentage range among the top K-ranked percentage values, we compare the \textit{consecutiveness} of $\mathrm{L}_{0}$ and \frameworkname, which is measured by the proportion of sentences with the entire top K choices being able to constitute a single consecutive range. For example e.g. \{\textit{10\%, 20\%, 30\%}\} constitutes a consecutive range from 10\% to 30\% while \{\textit{10\%, 30\%, 50\%}\} does not. Consecutiveness is based on the assumption that consecutive ranking of percentage values indicates better quantifier understanding as the semantic meaning of quantifier words does not leap between disjoint percentage values. To enlarge the possible ranges, we start by $\text{K}=3$ and include the results in Figure~\ref{fig: consecutiveness}, where \frameworkname has higher consecutiveness than $\mathrm{L}_{0}$, showing that \frameworkname has more consistent percentage inference behavior. Moreover, We select the primary scope by finding the consecutive scope (e.g. \{\textit{10\%-30\%}\} and \{\textit{10\%, 30\%, 50\%\}} in the previous example) of the largest aggregated $\mathrm{L}(p|q)$ among all consecutive scopes.

\begin{table*}[!ht]
    \small
    \centering
    \begin{minipage}[c]{1\textwidth}
        \centering
        \begin{tabular}{p{1.8cm}|c|c|c|c|c|c|c|c|c|c|c|c}
        \toprule
            \multirow{2}{*}{\Thead{Specificity}}  & \multicolumn{3}{|c}{\Thead{HIT@1}$\uparrow$} & \multicolumn{3}{|c}{\Thead{MRR}$\uparrow$} & \multicolumn{3}{|c}{\Thead{CrossEntropy}$\downarrow$} & \multicolumn{3}{|c}{\Thead{F1@\{1, 5\}}$\uparrow$} \\
            \cmidrule(lr){2-13}
            & Rnd. & $\mathrm{L}_{0}$ & $\mathrm{L}_{1}$ & Rnd. & $\mathrm{L}_{0}$ & $\mathrm{L}_{1}$ & Rnd. & $\mathrm{L}_{0}$ & $\mathrm{L}_{1}$ & Rnd. & $\mathrm{L}_{0}$ & $\mathrm{L}_{1}$\\
            \midrule
            Fully & 4.1 & 27.3 & \textbf{29.7} & 12.3 & 22.1 & \textbf{24.3} & 6.44 & \textbf{5.64} & 5.74 & 2.8/8.6 & 19.5/24.3 & \textbf{21.5}/\textbf{26.5} \\
            Partial & 8.2 & 26.4 & \textbf{28.5} & 11.6 & 21.2 & \textbf{21.7} & 7.78 &  \textbf{6.99} & 7.06 & 4.3/8.3 & 16.9/25.9 & \textbf{18.3}/\textbf{27.3} \\
            Indeterminable & 9.7 & 21.4 & 21.4 & 12.5 & 18.1 & \textbf{22.7} & 7.76 & 7.20 & \textbf{6.69} & 5.3/10.1 & \textbf{14.9}/18.2 & 14.8/\textbf{25.6} \\
            \midrule
            Total & 7.9 & 24.0 & \textbf{25.1} & 11.8 & 19.8 & \textbf{22.7} & 7.47 & 6.86 & \textbf{6.78} & 4.4/9.3 & 16.3/21.7 & \textbf{17.1}/\textbf{26.3}\\
            \bottomrule
        \end{tabular}
    \end{minipage}
    \caption{Performance of \frameworkname{} ($\mathrm{L}_{1}$) versus $\mathrm{L}_{0}$ on \datasetname{} using RoBERTa-large. Metrics are displayed on a 0-100 scale except for cross-entropy. \textit{Rnd.} is a random baseline (averaged over 5 seeds) where $\mathrm{L}(p|q)$ is sampled from $\mathcal{N}(0, 1)$ and normalized with softmax. The best results are bolded. The results show that the random baseline is worse than both $\mathrm{L}_{0}$ and \frameworkname in most metrics. \frameworkname performs better than $\mathrm{L}_{0}$ on almost all specificity levels and metrics.}
    \label{tab: QuRe result}
\end{table*}

We additionally compare the primary scopes between $\mathrm{L}_{0}$ and \frameworkname, through human preferences. For each quantifier word, we randomly sample 10 sentences where the top K inferences between $\mathrm{L}_{0}$ and \frameworkname differ for the same $\Tilde{S}_{q}$ with $\text{K}=5$. We then recruit 40 annotators from Amazon Mechanical Turk to select the more reasonable primary scope between $\mathrm{L}_{0}$ and \frameworkname given $\Tilde{S}_{q}$. We displayed the primary scopes for each $\Tilde{S}_{q}$ in random order to avoid biases. Examples are included in Table~\ref{tab:listener preference example} where \frameworkname generates smaller primary scopes for universal quantifiers like \textit{No} and \textit{All}, and larger primary scopes of other quantifiers which incorporate more vagueness. We leave the more general analysis in Appendix~\ref{sec: Human Preference of Examples in HVD}.

Table~\ref{tab: QuRe result} provides a comparison of the top percentage predictions with the gold scopes from \frameworkname and $\mathrm{L}_{0}$ in \datasetname. We observe that, in general, \frameworkname outperforms $\mathrm{L}{0}$ in several aspects. Firstly, the topmost prediction value from \frameworkname appears more frequently within the gold scope, leading to a higher HIT@1 score. Additionally, the percentage values within the gold scope are ranked higher among the top predictions by \frameworkname, resulting in a higher Mean Reciprocal Rank (MRR). Furthermore, there is a larger overlap between the primary scopes of \frameworkname and the gold scopes, as indicated by a higher F1 score. Moreover, \frameworkname predicts better primary scopes on a distance-based metric designed to measure the distance between scopes, and we include the results in Appendix~\ref{sec: metrics of top K percentages}. This finding aligns with the conclusion of \citet{carcassi-szymanik-2021-vs}, which suggests that listeners aim to minimize the distance from the speaker's observation when the communication success is measured based on distance.

\paragraph{Qualitative Analysis.} Examining examples generated by \frameworkname, we make several interesting observations. For fully determinable contexts, such as ``... only (2 out of the total 27) few school children were recognized..." with a gold scope of \textit{5\%-10\%} (the true rate was 2 children out of 27 = 7\%), \frameworkname provides a more accurate primary scope. In this case, \frameworkname predicted a scope of \textit{0\%-5\%}, while $\mathrm{L}_{0}$ predicted a scope of \textit{0\%}. For partially determinable contexts, such as ``... calculating from less than few ... to 13\%..." (indicating a partially determinable percentage scope of less than 13\%), with a gold scope of \textit{5\%-10\%}, \frameworkname often generates a broader scope than $\mathrm{L}_{0}$. In this case, \frameworkname predicts \textit{0\%-15\%}, which is more expansive than $\mathrm{L}_{0}$'s prediction of \textit{10\%-15\%}. For some indeterminable sentences like ``... its alcohol content usually is very little." with a gold scope of \textit{0\%-5\%}, \frameworkname provides a primary scope of \textit{0\%-5\%}, while $\mathrm{L}_{0}$ predicts a significantly distant scope of \textit{60\%-70\%}. Appendix~\ref{sec: QuRe examples with PRESQUE extended} provides a more comprehensive set of examples.

\section{Conclusion}
Generalized quantifiers are widely used for addressing satisfaction in natural language. However, the quantifier understanding abilities of foundation models are not well-studied or well-supported by the existing benchmarks based on factual context. In this work, we study quantifier understanding by proposing the \frameworkname framework that formulates the problem in a pragmatics reasoning perspective and the format of NLI. And we collect a dataset \datasetname that includes human annotated quantifiers and percentage mentions for Wikipedia sentences. From the experimental results on the HVD dataset and our collected \datasetname dataset, the \frameworkname achieves better performance compared to the direct interpretation of quantifier semantics by a literal listener model.

\section*{Acknowledgments}
The authors would like to thank the anonymous reviewers for their suggestions and feedback on the work. This work was supported in part by NSF grant DRL2112635. The views contained in this article are those of the authors and not of the funding agency. We also thank Daniel Fried for useful discussion and suggestions to this work.

\section*{Limitations}
In this work, we investigate the quantifier understanding abilities of several foundation models and collect a dataset \datasetname that we expect will substantially benefit research on quantifier semantics. However, despite the value of our dataset and the promising results from the \frameworkname framework, our analysis and findings have some notable limitations. First, we note that our study and dataset still focus on a small part of the generalized quantifiers and likely do not cover the entire spectrum of quantifier semantics. Second, the sentences in our dataset all come from Wikipedia. Consequently, the performance of \frameworkname and the generalizability of our findings to other domains or languages remains an open question. Finally, assigning precise percentage scopes to quantifiers can be a challenging or even impossible task, since quantifier semantics are complex and depend on many factors beyond those analyzed here. In particular, these may subjectively depend on the domain, an annotator's background of knowledge or culture, comfort with the mathematics of percentages, and Bayesian vs Frequentist interpretations of percentage numbers, among many other factors. Thus, ambiguities and subjectivity are natural when determining a quantifier's scope. Our dataset and analysis largely skirt many of these complex issues.

\section*{Ethics and Broader Impact}
We employ crowdsourcing through Amazon Mechanical Turk for (a) certain annotations of our dataset, \datasetname, (b) understanding human interpretation of quantifier semantics, and (c) human evaluation of \frameworkname and baselines. In all our crowdsourcing tasks we do not collect any personally identifiable information about the turkers and all the turkers were paid above minimum wage, which is included in Appendix~\ref{sec: annotation recruitment}. We released our crowdsourcing tasks to turkers in the USA and constrained the approval rate of the turkers to be above 98\% to ensure good-faith turkers.

Besides, the prevalence of quantifiers in naturally occurring corpora would inherit the generation behavior of models. \frameworkname, as one step towards revealing the quantifier understanding ability of foundation models, could be helpful in more accurately interpreting the meaning in model-generated text. It could also support automatic pipelines for knowledge-intensive reasoning that include quantifications, or logical reasoning in natural language.

\bibliography{emnlp2023}
\bibliographystyle{acl_natbib}

\appendix

\section{Example Metadata of \datasetname}
\label{sec: QuRe metadata example}
\begin{figure}[!ht]
    \centering
    \includegraphics[width=1\linewidth]{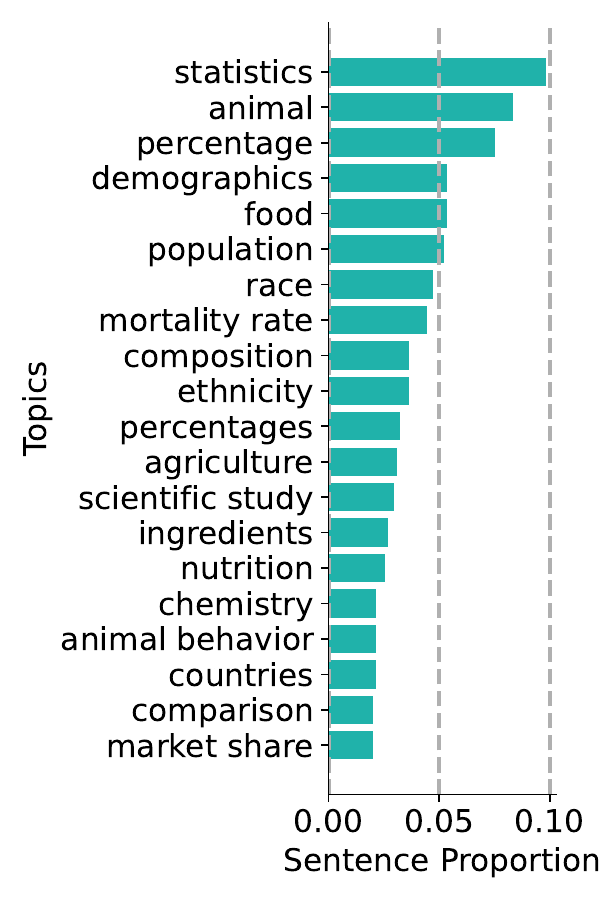}
    \caption{Top 20 sentence topic statistics in \datasetname. The most frequent sentence topics, where nearly 10\% sentences include the topic \textit{statistics} and 8\% sentences include \textit{animal}. Topics like \textit{demographics}, \textit{food} and \textit{population} also cover more than 5\% sentences in \datasetname.}
    \label{fig: sentence topic statistics}
\end{figure}

To investigate the topic coverage of \datasetname sentences, we use GPT-3.5-turbo to generate 3 topics for each sentence, using instruction in Appendix~\ref{sec: sentence topic generation}. The most frequent topics are listed in Figure~\ref{fig: sentence topic statistics}, where nearly 10\% sentences are about \textit{statistics}, followed by \textit{animal}, \textit{percentage} and \textit{demographics}.

\begin{table*}[]
    \small
    \centering
    \begin{tabular}{p{6.5cm}|p{6.5cm}|p{1.8cm}}
    \toprule
        \Thead{[Wiki entity] Original Sentences} & \Thead{[Specificity, Expression] 
  \datasetname{} Sentences}& \Thead{Topics}\\
        \midrule
        \small{\textbf{[Human]} Most humans (61\%) live in Asia; the remainder live in the Americas (14\%), Africa (14\%), Europe (\underline{11\%}), and Oceania (0.5\%).Within the last century, humans have explored challenging environments such as Antarctica, the deep sea, and outer space.} & \small{\textbf{[Fully, $0.11$]} Most humans (61\%) live in Asia; the remainder live in the Americas (14\%), Africa (14\%), \underline{some} Europe, and Oceania (0.5\%).Within the last century, humans have explored challenging environments such as Antarctica, the deep sea, and outer space.} & population\newline continents\newline exploration\\
        \midrule
        \small{\textbf{[The Jungle Book (2016 film)]} The Jungle Book was shown across 4,028 theaters of which 3,100 theaters (\underline{75\%}) were in 3D, including 376 IMAX screens, 463 premium large format screens, and 145 D-Box locations.} & \small{\textbf{[Fully, $0.75$]} The Jungle Book was shown across 4,028 theaters of which \underline{most} (3,100) theaters were in 3D, including 376 IMAX screens, 463 premium large format screens, and 145 D-Box locations.} & theaters\newline movie release\newline 3D technology\\
        \midrule
        \small{\textbf{[Electric car use by country]} The EV market share of total new and used cars first registered during 2018 was \underline{2.8\%} based on 5,557 out of a total of 198,600 first registered cars.7,542 vehicles were registered in this country over 2019.} & \small{\textbf{[Fully, $0.028$]} The EV market share of total new and used cars first registered during 2018 was \underline{small} based on 5,557 out of a total of 198,600 first registered cars. 7,542 vehicles were registered in this country in 2019.} & electric vehicles\newline market share\newline registration numbers\\
        \midrule
        \small{\textbf{[List of blade materials]} Prior to 2002, INFI contained \underline{0.5\%} Carbon, 0.74\% Nitrogen, about 1\% Cobalt, and about 0.1\% Nickel.} & \small{\textbf{[Partially, $0.005$]} Prior to 2002, INFI contained \underline{tiny} levels of Carbon, 0.74\% Nitrogen, about 1\% Cobalt, and about 0.1\% Nickel.} & chemical composition\newline INFI\newline elements\\
        \midrule
        \small{\textbf{[Housing in the United Kingdom]} British dwellings had the oldest age profile in the EU with over 60\% being built before 1960, and with only just over \underline{10\%} being built between 1991-2010.} & \small{\textbf{[Partially, $>0.1$]} British dwellings had the oldest age profile in the EU with over 60\% being built before 1960, and with \underline{some} being built between 1991–2010.} & age\newline housing statistics\newline construction date\\
        \midrule
        \small{\textbf{[Ice cream]} Gelato typically contains 7-8\% fat, less than ice cream's \underline{minimum of 10\%.}} & \small{\textbf{[Partially, $>=0.1$]} Gelato typically contains 7–8\% fat, less than the \underline{moderate} amount found in ice cream.} & food\newline comparison\newline fat percentage\\ 
        \midrule
        \small{\textbf{[Tobacco]} A study published in Morbidity and Mortality Weekly Report found that in 2019 approximately one in four youths (\underline{23.0\%}) in the U.S. had used a tobacco product during the past 30 days.} & \small{\textbf{[Partially, $0.23$]} A study published in Morbidity and Mortality Weekly Report found that in 2019,  \underline{some} (one in four) youths in the U.S. had used a tobacco product during the past 30 days.} & youth\newline tobacco use\newline scientific study\\
        \midrule
        \small{\textbf{[Polish cuisine]} It is typically made from rye bread, usually known as black bread, and is not classified as an alcoholic beverage in Poland, as its alcohol content usually ranges from \underline{0\% to 2\%.}} & \small{\textbf{[Indeterminable, $0-0.02$]} It is typically made from rye bread, usually known as black bread, and is not classified as an alcoholic beverage in Poland, as its alcohol content usually is very \underline{little}.} & food\newline beverage\newline alcohol content\\           
        \midrule
        \small{\textbf{[List of blade materials]} In order for a steel to be considered stainless it must have a Chromium content of \underline{at least 10.5\%.}} & \small{\textbf{[Indeterminable, $>=0.105$]} In order for a steel to be considered stainless it must have \underline{some} Chromium content.} & steel\newline metallurgy\newline composition\\      
        \midrule
        \small{\textbf{[British military bands]} The average age of the 304 drummers at Waterloo was 25, with \underline{about 10\%} being boys under 16.} & \small{\textbf{[Indeterminable, $\sim0.1$]} The average age of the 304 drummers at Waterloo was 25, with \underline{some} being boys under 16.} & age\newline music\newline statistics\\   
        \bottomrule
    \end{tabular}
    \caption{Example data of \datasetname, with target percentage mention and quantification \underline{underlined}. The header marks either the Wikipedia entity where the sentence is extracted or the specificity and the generated percentage scopes. For example, for the Jungle Book sentence, the percentage scope 75\% can be fully specified by the proportion of 3100 over 4028 theatres, while for the sentence about Gelato, the content before the percentage mention indicates that the fat content of ice cream is higher than 7-8\%, but cannot provide more information to further narrow down the scope, and therefore the specificity is \textit{partially}.}
    \label{tab: QuRe dataset examples}
\end{table*}

\section{\frameworkname Examples in \datasetname}
\label{sec: QuRe examples with PRESQUE extended}
We include several examples in Table~\ref{tab: QuRe examples shortened} where $\Tilde{S}_{q}$ as well as the specificity level, and the $\Tilde{S}_{p}$ as well as the golden percentage scope are located on the upper and lower half of each block. We can see that \frameworkname provides more accurate primary scopes for fully determinable sentences.
\begin{table*}[]
    \small
    \centering
    \begin{tabular}{p{7.8cm}|l|c|c|c}
    \toprule
        \Thead{[GS.] $\text{Sentence}_{\text{Q}}$} / \Thead{[SPC.] $\text{Sentence}_{\text{P}}$} & \Thead{Primary Scope} & \Thead{MRR}  & \Thead{F1@5} & \Thead{CE}\\
        \midrule
        \small{[\textbf{F}] In 57 separate fights, one loss was observed to Neope goschkevitschii, giving V. mandarinia a \underline{large} winning rate.} & $\mathrm{L}_{0}$: 5\%-20\% & 0.11 & 0.00 & 7.67\\
        \cmidrule(lr){2-5}
        \small{[\textbf{95\%-100\%}] In 57 separate fights, one loss was observed to Neope goschkevitschii, giving V. mandarinia a win rate of \underline{98.3\%}.} &  $\mathrm{L}_{1}$: 85\%-100\% & \textbf{0.67} & \textbf{0.67} & \textbf{3.52}\\
        \midrule
        \small{[\textbf{F}] In the 2017 Dutch study, only (2 out of the total 27) \underline{few} school children recognized that the website was a hoax.} & $\mathrm{L}_{0}$: 0\% & 0.08 & 0.00 & 7.79\\
        \cmidrule(lr){2-5}
        \small{[\textbf{5\%-10\%}] In the 2017 Dutch study only 2 out of the total 27 school children \underline{(7\%)} recognized that the website was a hoax.} &  $\mathrm{L}_{1}$: 0\%-5\% & \textbf{0.11} & \textbf{0.50} & \textbf{6.36}\\
        \midrule
        \small{[\textbf{P}] From 4 locations in different parts of Europe, a \underline{large} number had clutch size of 2, 41\% had size of 3, clutches of 1 and 4 each constituted about 8\%.}& $\mathrm{L}_{0}$: 30\%-40\% & 0.22 & 0.40 & 6.29\\
        \cmidrule(lr){2-5}
        \small{[\textbf{40\%-45\%}] From 4 locations in different parts of Europe, \underline{43\%} had clutch size of 2, 41\% had size of 3, clutches of 1 and 4 each constituted about 8\%.} & $\mathrm{L}_{1}$: 30\%-45\% & \textbf{0.33} & \textbf{0.67} & \textbf{4.92}\\
        \midrule
        \small{[\textbf{P}] The empirical occurrence of regenerated claws in fishery harvests is low, with studies on stone crabs calculating from less than \underline{few} (Davis et al., 1978), to 13\% (Florida Fish and Wildlife Conservation Commission, 2011).} & $\mathrm{L}_{0}$: 10\%-15\% & 0.17 & 0.50 & 7.79\\
        \cmidrule(lr){2-5}
        \small{[\textbf{5\%-10\%}] The empirical occurrence of regenerated claws in fishery harvests is low, with studies on stone crabs calculating from less than \underline{10\%} (Davis et al., 1978), to 13\% (Florida Fish and Wildlife Conservation Commission, 2011).} &  $\mathrm{L}_{1}$: 0\%-15\% & \textbf{0.50} & \textbf{0.67} & \textbf{4.40}\\
        \midrule
        \small{[\textbf{I}] It is typically made from rye bread, usually known as black bread, and is not classified as an alcoholic beverage in Poland, as its alcohol content usually is very \underline{little.}} & $\mathrm{L}_{0}$: 60\%-70\% & 0.06 & 0.00 & 6.97\\
        \cmidrule(lr){2-5}
        \small{[\textbf{0-5\%}] It is typically made from rye bread, usually known as black bread, and is not classified as an alcoholic beverage in Poland, as its alcohol content usually ranges from \underline{0\% to 2\%}.} &  $\mathrm{L}_{1}$: 0\%-5\% & \textbf{0.33} & \textbf{1.00} & \textbf{4.16}\\
        \midrule
        \small{[\textbf{I}] Chlamydospore germination requires 30 to 52 hours, with a \underline{moderate} germination success rate. Spore production is highest at midday, relative to temperature increase and relative humidity decrease.} & $\mathrm{L}_{0}$: 30\%-35\% & 0.13 & 0.50 & 18.85\\
        \cmidrule(lr){2-5}
        \small{[\textbf{30\%-55\%}]Chlamydospore germination requires 30 to 52 hours, with a germination success rate of \underline{32 to 54\%}. Spore production is highest at midday, relative to temperature increase and relative humidity decrease.} &  $\mathrm{L}_{1}$: 40\%-50\% & \textbf{0.22} & \textbf{0.67} & \textbf{16.17}\\
        \bottomrule
    \end{tabular}
    \caption{Examples of \frameworkname{} ($\mathrm{L}_{1}$) versus $\mathrm{L}_{0}$. The sentences are paired with percentages and the corresponding sentence with quantifiers, with the target percentage and quantification phrase \underline{underlined}. The headings mark either the gold scope (\textbf{GS}) or the specificity levels (\textbf{SPC.}) with \textbf{[F/P/I]} being fully/partially/indeterminable respectively. \textbf{CE} stands for cross-entropy. Bolded figures are better results. Predictions are collected from $\mathcal{W}_{\beta=0.05}$. $L_{1}$ achieves better MRR and cross entropy then $L_{0}$ among different sentence inferring categories.
    }
    \label{tab: QuRe examples shortened}
\end{table*}

\begin{table*}[]
    \centering
    \begin{tabular}{l|l|l|p{0.35\linewidth}}
    \toprule
        \Thead{concept} & \Thead{feature} & \Thead{Annotations} & \Thead{Sentence based on template}\\
        \midrule
        rock & has\_minerals & all, all, most & All rocks have minerals.\\
        van & has\_sliding\_doors & most, most, most & Most vans have sliding doors.\\
        sandpaper & has\_fine\_sand\_covering\_it & some, some, all & Some sandpapers have fine sand covering it.\\
        banana & is\_round & no, no, no & No bananas are round.\\
        tricycle & used\_for\_transportation & all, few, few & Few tricycles are used for transportation.\\
        \bottomrule
    \end{tabular}
    \caption{Sample $\langle$concept, feature$\rangle$, human annotations for the quantifiers, and the corresponding HVD sentences that serve as $\Tilde{S}_{q}$ using the majority quantifier annotation.}
    \label{tab:HVD examples}
\end{table*}

\section{Discussion}
\label{sec: quantifer world vs numerical world}
\begin{figure*}[!ht]
    \centering
    \includegraphics[width=1\linewidth]{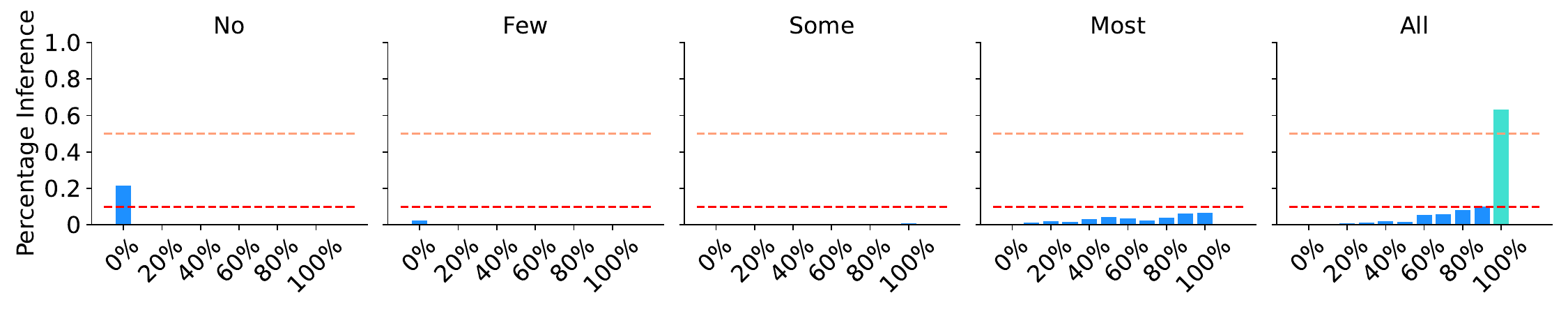}
    \includegraphics[width=1
    \linewidth]{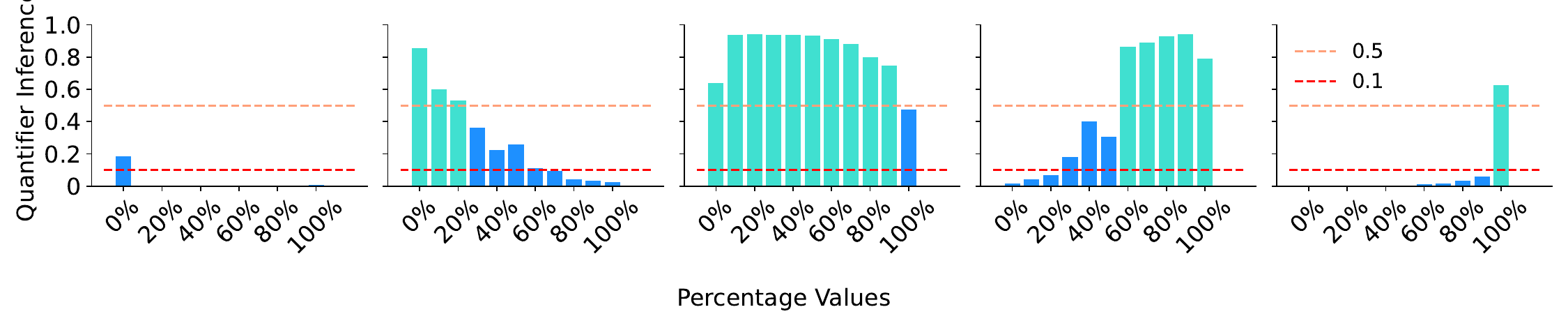}
    \caption{Comparison of the NLI model conducting percentage inference ($\text{Entailment}(\Tilde{S}_{q}, \Tilde{S}_{p})$, upper) and quantifier inference ($\text{Entailment}(\Tilde{S}_{p}, \Tilde{S}_{q})$, lower) in HVD. The NLI model is more sensitive in quantifier inference in general. Cyan bars indicate values higher than the upper threshold (0.5). Note that the bar values do not stand for probabilities and do not sum up to 1.}
    \label{fig: literal speaker numerical world}
\end{figure*}

The semantic understanding of a pragmatic listener is proportional to the product of two entailment-based probabilities, where the premise and hypothesis are flipped with respect to each other (i.e., the premise used for $S_0(q|p)$ is the hypothesis for calculating $P(p|q)$ in the prior (Eq.~\ref{eq:bayesian prior}). To arrive at an intuitive understanding of why considering the flipped premise-hypothesis pairs, we analyze the sensitivity of NLI models (specifically, RoBERTa fine-tuned NLI model) towards percentage values (quantifier inference) and quantifier words (percentage inference) in premises.
The distribution of average entailment scores over all the premises is shown in Figure~\ref{fig: literal speaker numerical world}. The upper part of the figure shows the result of percentage inference and the lower part shows the result of quantifier inference, with two thresholds, 0.1 and 0.5. We can observe that the NLI model is more sensitive to the percentage values in the premise than quantifiers. The entailment scores of percentage inference is relatively low, which is led by high neutral scores, making it challenging to identify the percentage scope for each quantifier. For example, \textit{few}, \textit{some} and \textit{most} don't have any percentage values that exceed even the lower threshold. The lower half of Figure~\ref{fig: literal speaker numerical world}, however, demonstrates more interpretable entailment distribution, where the percentage scopes of \textit{few}, \textit{some} and \textit{most} can be interpreted as \textit{0\%-20\%}, \textit{0\%-90\%} and \textit{60\%-100\%} by the higher threshold. In short, NLI models are more sensitive to interpreting accurate numerical premises, which has also been observed that NLI performs better with accurate premises~\citep{thukral-etal-2021-probing, Richardson2019} where premises with quantifiers are less accurate than premises with percentage values.

\begin{figure}[!ht]
    \centering
    \includegraphics[width=1\linewidth]{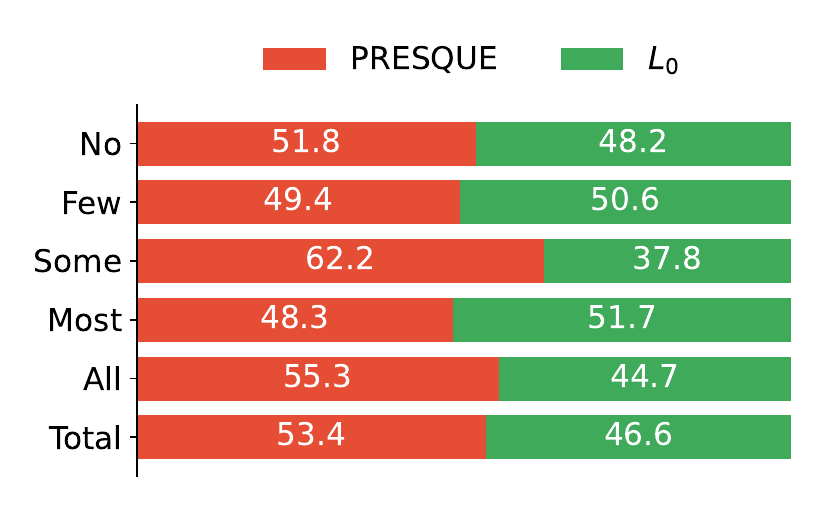}
    \caption{Listener preference from humans for HVD examples. The inference from \frameworkname is in general, preferred by humans than $\mathrm{L}_{0}$, while the preference of each quantifier differs (\frameworkname is more preferred for \textit{No, Some} and \textit{All}). 
    }
    \label{fig: listener preference}
\end{figure}

\section{Annotator Recruitment}
\label{sec: annotation recruitment}

We use Amazon Mechanical Turk as the crowdsourcing platform for different annotation aspects of \datasetname as described in previous paragraphs. To finalize our pool of turkers, we released a qualification task to test basic understanding of quantifier semantics. The annotators are selected to base on the United States, completed more than 1000 HITs with more than 98\% approval rate. For annotator that participates the final \datasetname collection in Section~\ref{sec: dataset collection}, they can make at most one annotation mistake in sentence rephrasing of the qualification task. The qualification task had a pass rate of 38\% and we recruited 18 turkers for the main annotation tasks of \datasetname. Annotators are paid about \$ 7.30/hr on average. Besides, annotators recruited for human interpretation of quantifiers are paid about \$ 9/hr on average. And the annotators recruited for human preference of percentage scopes are paid about \$ 9.60/hr on average.

\section{Percentage Scope Generation Details}
\label{sec: percentage scope grounding details}
With granularity $g$, window size $w$, the grounded percentage scope can be determined in Table~\ref{tab: percentage scope grounding}.
\begin{table}[ht]
    \small
    \centering
    \begin{tabular}{c|l}
    \toprule
        \Thead{Expression} & \Thead{Percentage Scope}\\
        \midrule
        $p$ & $p$\\
        \midrule
        $>p$  & $(p, p + w \cdot g]$\\
        \midrule
        $>=p$  & $[p, p + w \cdot g]$\\
        \midrule
        $<p$  & $[p - w \cdot g, p)$\\
        \midrule
        $<=p$  & $[p - w \cdot g, p]$\\
        \midrule
        $p_{1}-p_{2}$  & $[p_{1}, p_{2}]$\\
        \midrule
        $\sim p$ & $[p - w \cdot g, p + w \cdot g]$\\
        \bottomrule
    \end{tabular}
    \caption{Percentage scope grounding rules with granularity $g$, window size for approximation $w$. And $[p_{\text{min}}, p_{\text{max}}]$ is the smallest scope in $\mathcal{W}_{\beta}$ that includes the above scope. The scope would be cut off at 0 and 1.}
    \label{tab: percentage scope grounding}
\end{table}

For HVD, $\beta$ is set to be 0.1. And in experiments on \datasetname, $\beta$, $w$, $g$ are set to be 0.05, 2, 0.01 respectively unless specified.

\section{Human Preference of HVD Examples}
\label{sec: Human Preference of Examples in HVD}
Figure~\ref{fig: listener preference} shows \frameworkname is in general preferred over $\mathrm{L}_{0}$  by the annotators, while the preferences may differ for different quantifiers. The primary scopes of \frameworkname for \textit{no}, \textit{some} and \textit{all} are more preferred than $\mathrm{L}_{0}$ by the annotators. \textit{Some} and \textit{all} have $p < 0.05$ in chi-squared test.

\section{Distance-based Scope Evaluation}
\label{sec: metrics of top K percentages}
To measure the primary scope of $L(p|q)$ and the gold scope $[p_{\text{min}}, p_{\text{max}}]$ in distance-based metrics. We compute the minimal scope distance (MSD) over the top K predictions (MSD@K). Specifically,\\
\begin{small}
\begin{flalign}
    \small
    &\text{MSD@K} = \sum_{p' \in \text{TopK p}} \frac{\mathbb{I}[p' \notin s_{\text{golden}}]}{\text{B}_{m}}\text{min}(p_{\text{min}} - p', p' - p_{\text{max}}) && \nonumber\\
    &\text{where} \quad \text{B}_{m} =  p_{\text{max}} - p_{\text{min}} + \beta, \quad p' \in \mathcal{W}_{\beta} \nonumber
\end{flalign}
\end{small}

\begin{figure*}[!ht]
    \centering
    \includegraphics[width=1\linewidth]{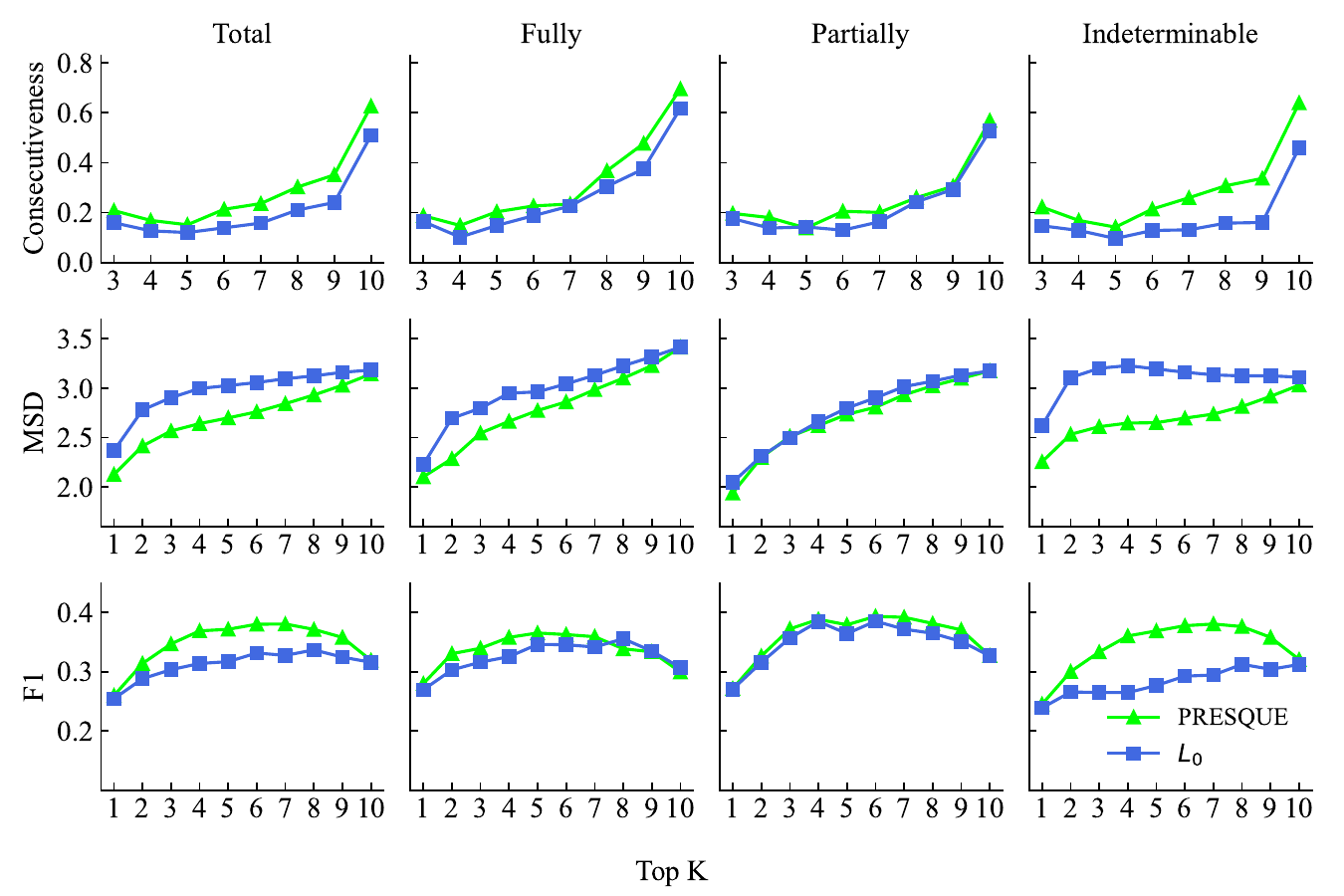}
    \caption{Consecutiveness ($\uparrow$), MSD ($\downarrow$) and F1 ($\uparrow$) of \frameworkname (lime) and $\mathrm{L}_{0}$ (blue) on \datasetname with $\beta=0.1$ for better illustration. \frameworkname has higher consecutiveness, lower MSD and higher F1 than $\mathrm{L}_{0}$ across all specificity levels. 
    }
    \label{fig: consecutiveness-intervalDistance}
\end{figure*}

\section{Instruction for Sentence Filtering}
\label{sec: instruction-quantifier-filtering}
\fbox{
\begin{minipage}{18em}
\small
In this task, you will determine whether a given sentence that has one or more quantifier values mentioned can have those quantifier values replaced by a natural language quantifier like 'some', 'most' or 'generally'. \\

Example sentences that meet the criteria are like 'It consists of about 80\% water, soluble minerals (nearly 3\% with half of the potassium) and polyphenols.' and sentences that don't meet the requirement are like '180.1 million were rides on SEPTA's 'city transit' network. Ridership had decreased 13\% from 2014 to 2019 due to many factors.' where the percentage value represents incremental percentage changes or comparisons (e.g. 'drop by 50\%' or '20\% higher', 'X\% better') instead of absolute percentages. \\

Do you think the following sentence meets the requirement? Answer in Yes or No:
\end{minipage}}

\section{Instruction for Percentage Scope Generation}
\label{sec: instruction-mathematical expression}
The instruction for percentage scope generation is shown below.\\

\fbox{
\begin{minipage}{18em}
\small
In this task, you will give a sentence and a quantifier expression in the sentence, and you need to convert that quantifier expression into a mathematical expression. For example, for the sentence 'About 30\% of homes are owned outright by their occupants. [30\%]', you are given 30\% in the bracket, and the corresponding mathematical expression is ~0.3, where ~ means approximate. Similarly, all the other mathematical operations supported include > meaning 'more than' (e.g. 'more than 80\%' would be > 0.8), < meaning 'less than', '-' meaning range (e.g. '20\% to 50\%' would be 0.2-0.5) and null meaning exact (e.g. 'takes up 20\%' would be 0.2). Answering the expression itself is enough, don't repeat the sentence or use additional English words other than the operations. Also, try to avoid using '<' and '>' if you can formulate the range by using '-'. Now, please do the same conversion for the following sentence: 
\end{minipage}}

\section{Instruction for Sentence Topic Generation}
\label{sec: sentence topic generation}
\fbox{
\begin{minipage}{18em}
\small
Please use three or four labels to categorize a given sentence (starts with "sentence"), including the topics of the contents, split with semicolons.

For example,

Sentence: In fact, a 2006 survey found that trapping as a solution to beaver problems had a 79\% failure rate within two years due to resettlement by new beavers.
Labels: scientific study; animal; rate

Sentence: Among individual countries, the proportion of urban residents living in slum areas in 2009 was highest in the Central African Republic (95.9\%), Chad (89.3\%), Niger (81.7\%), and Mozambique (80.5\%). The distribution of slums within a city varies throughout the world.
Labels: population; ranking; countries.

Now, please label the following sentence:
\end{minipage}}

\section{Instruction for Human Evaluation}
\label{sec: human evaluation instruction}
The instruction for collecting human perception of quantifier words in Section~\ref{sec: human evaluation L1} is displayed as\\

\fbox{\begin{minipage}{18em}
\small
This form contains several natural language quantifiers. The users are expected to pick one/two numerical percentages from the provided list of percentages such that most accurately bound the range of the given quantifier to the best of his/her knowledge and online searching is not encouraged.\\

Users can use different (real or imaginative) statements as examples to help estimate the range, such as `No water comes from the sky.' and `Most sea cucumbers are scavengers.'.\\

The users can select no more than 2 options to mark the lower and upper bound of the range, if they believe only one percentage would apply, they can select only 1 option.\\

An example of `All' stands for' with a statement is `All sugars are white.' The users are expected to select `100\%' or a range (based on the user's understanding) from all provided percentages as the range for `All', and the paraphrase becomes `A\% to B\% sugars are white.' (A and B are the selections and can be the same) which becomes the most appropriate paraphrase of `All sugars are white'. Note that the statement itself does not necessarily involve factuality (in fact, sugars can have various colors).
\end{minipage}}\\

The instruction for collecting listener preference of $\mathrm{L}_{0}$ and $\mathrm{L}_{1}$ in Section~\ref{sec: model selection} is displayed as\\

\fbox{\begin{minipage}{18em}
\small
This form contains several statements (e.g. sugars are white) with natural language quantifiers (e.g. all). The users are expected to pick the more appropriate percentage range from the provided two options such that accurately bound the range of the given quantifier to the best of his/her knowledge and online searching is not encouraged.\\

An example statement with quantifier is 'All sugars are white.', and two example options are `0\%-30\%' and `90\%-100\%'.\\

In this example, the users are expected to select `90\%-100\%', which results in that `90\%-100\% sugars are white.' better describes `All sugars are white.'. Note that the statement itself does not necessarily involve factuality (in fact, sugars can have various colors).
\end{minipage}}

\section{Quantifier Understanding of GPT-3.5-turbo}
\label{sec: quantifier understanding gpt-3.5}

Although we mainly focus on NLI models to develop \frameworkname, we also test the performance of \datasetname on GPT-3.5-turbo using the following instruction.\\

\fbox{
\begin{minipage}{18em}
\small
In this task, you are given a sentence (starts with `Sentence:') containing a predicate with a quantifier, and you need to provide a percentage scope that the predicate satisfies. \\

For example, if you are given “Sentence: Some apples are red.” for the quantifier `some', and you believe 37\%-42\% apples are red. Then the percentage scope for “some apples are red” would be 37\%-42\%.\\

The scope you can choose should be rounded in the granularity of 5
\%. In the previous `apples are red' example, your answer will be "35\%-45\%". Not that the percentage value cannot exceed 0\% and 100\%. You can also select one single percentage value for the scope.\\

Please provide a percentage scope for “some” in the following sentence.\\

Sentence: 
\end{minipage}}\\

In the example instruction shown above, the quantifier \textit{some} would be replaced by the target quantifier that appeared in the sentence attached to the instruction. For example, for sentence \textit{Adult clams can get most of their nutrients from the algae and the rest from filter feeding.} (gold scope 65\%-100\%), the output of GPT-turbo-3.5 for quantifier \textit{most} is \textit{60\%-80\%}.

Overall, GPT-3.5-turbo achieves 0.28 F1 score of the quantifier understanding task on \datasetname, under the same configuration in Appendix~\ref{sec: percentage scope grounding details}, which is slightly higher than the F1@5 performance of \frameworkname. However, we are aware that text-to-text models like GPT-3.5-turbo still suffer from hallucination and the output is unstable due to temperature-based sampling. Meanwhile, the \frameworkname in this work is agnostic to the backbone model choices and can be applied to any models that score the entailment relation between sentences.

\section{Annotation Task Interface}
\label{sec: annotation task interface}

\begin{figure*}[!ht]
    \centering
    \includegraphics[width=1\linewidth]{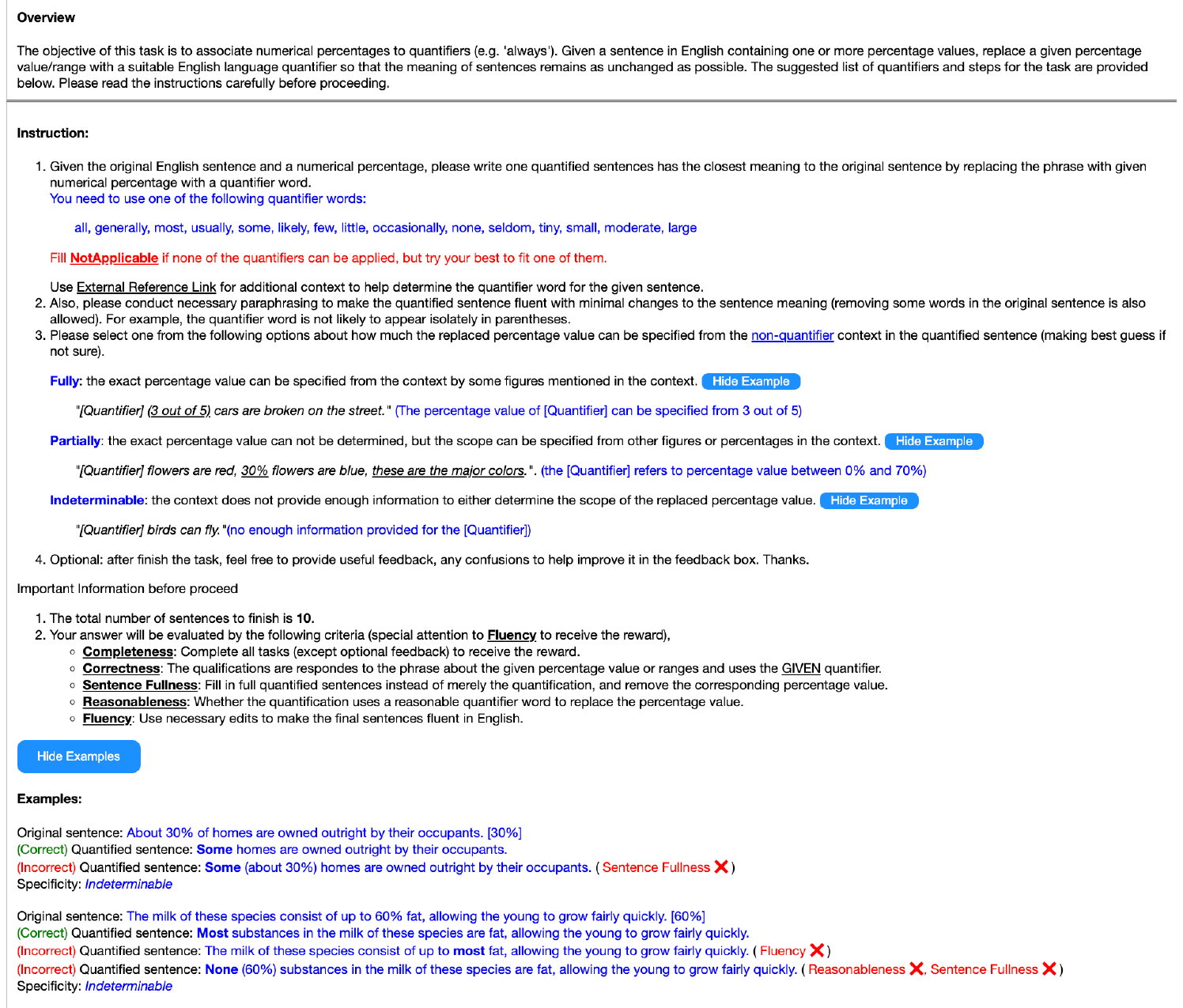}
    \caption{Instruction for the annotation task.}
    \label{fig: qualification task instruction}
\end{figure*}

\begin{figure*}[!ht]
    \centering
    \includegraphics[width=1\linewidth]{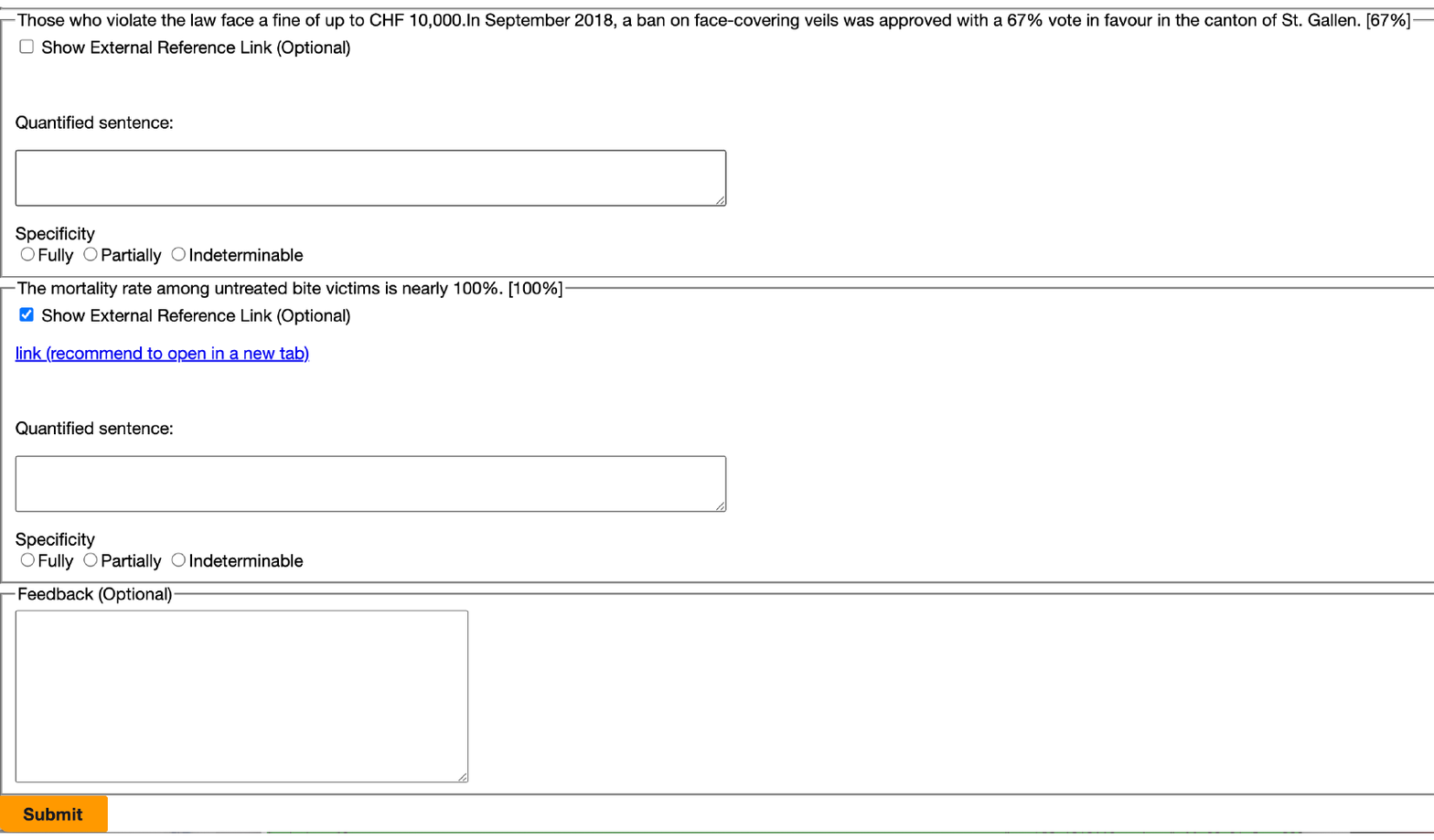}
    \caption{Interface of the example annotation task. Each sentence comes with a target percentage in the bracket at the end of the sentence that directs the target percentage mentioned in the sentence (e.g. \textit{100\%} for \textit{nearly 100\%} in the second sentence). If there are multiple target percentage mentions that share the percentage value, a positional indicator would be attached. Besides, an optional reference link is provided by checking the box.}
    \label{fig: qualification task example}
\end{figure*}

The instruction for the qualification task in collecting \datasetname is included in Figure~\ref{fig: qualification task instruction}, and Figure~\ref{fig: qualification task example} shows the example tasks annotators need to complete.

\end{document}